# "HOT" ChatGPT: The promise of ChatGPT in detecting and discriminating hateful, offensive, and toxic comments on social media

Lingyao Li[1]*, Lizhou Fan[1], Shubham Atreja[1], Libby Hemphill[1]*

[1]School of Information, University of Michigan, Ann Arbor, MI
*Corresponding author: lingyaol@umich.edu, libbyh@umich.edu

## Abstract

Harmful content is pervasive on social media, poisoning online communities and negatively impacting participation. A common approach to address this issue is to develop detection models that rely on human annotations. However, the tasks required to build such models expose annotators to harmful and offensive content and may require significant time and cost to complete. Generative AI models have the potential to understand and detect harmful content. To investigate this potential, we used ChatGPT and compared its performance with MTurker annotations for three frequently discussed concepts related to harmful content: Hateful, Offensive, and Toxic (HOT). We designed five prompts to interact with ChatGPT and conducted four experiments eliciting HOT classifications. Our results show that ChatGPT can achieve an accuracy of approximately 80% when compared to MTurker annotations. Specifically, the model displays a more consistent classification for non-HOT comments than HOT comments compared to human annotations. Our findings also suggest that ChatGPT classifications align with provided HOT definitions, but ChatGPT classifies "hateful" and "offensive" as subsets of "toxic". Moreover, the choice of prompts used to interact with ChatGPT impacts its performance. Based on these insights, our study provides several meaningful implications for employing ChatGPT to detect HOT content, particularly regarding the reliability and consistency of its performance, its understanding and reasoning of the HOT concept, and the impact of prompts on its performance. Overall, our study provides guidance about the potential of using generative AI models to moderate large volumes of user-generated content on social media.

## Keywords

# 1. Introduction

Harmful behaviors, such as cyberbullying, hate speech, and harassment, can create a hostile and unpleasant online environment, which may drive users away, reduce engagement, and cause harm to individuals (Fan et al., 2020; Whittaker & Kowalski, 2015). These negative effects make it crucial to develop models that can detect and moderate such content (Alorainy et al., 2019; Djuric et al., 2015). By using these models, platforms can take prompt action to remove harmful content, warn users, or provide resources to mitigate its impact. This not only safeguards users from harm but also fosters a safer and healthier online environment for all. Traditionally, one prevalent method for building these models involves human annotation of potentially toxic or hateful discourse (Davidson et al., 2017; Mathew et al., 2021), which can serve as an initial step in training algorithms to recognize and filter such content.

However, the process of annotating harmful content for model training poses several challenges. First, these tasks expose annotators to harmful and offensive content, such as violence, racism, sexism, and threats, which can have negative impacts on their mental health. Consequently, this limits the pool of available annotators and also restricts the amount of content they can reasonably review. Second, the financial costs of annotating data are high (Talat et al., 2021). Toxic content is relatively rare (Ibrahim et al., 2018), and therefore, a large number of documents must be annotated to obtain representation for effective model training. This can result in a considerable investment of time and resources. Third, the demographics of annotators can affect the objectivity of annotation tasks (Díaz et al., 2022). Due to variations in social cultural backgrounds of human annotators, the annotation results can be unstable even if they are normalized.

The challenges associated with manual annotation highlight the need for alternative approaches that can accurately detect and moderate online harm. Generative AI is one such approach that holds promise in this context (Huang et al., 2023). Generative AI models are often trained on large datasets of existing examples to identify common patterns and features of toxic content. Moreover, generative AI models have the potential to understand and respond in ways that mimic human conversations, which can be beneficial for detecting subtle forms of toxicity such as sarcasm or irony. As a result, there is a growing trend in exploring the potential of generative AI for performing a variety of tasks, including toxicity detection (Sorensen et al., 2022; Strobelt et al., 2022; Zhou et al., 2023).

While generative AI models, particularly the latest generative pre-training transformer (GPT) models for ChatGPT, have shown promise in detecting online toxicity (Huang et al., 2023), their full potential for classifying harmful content is not yet fully understood. In particular, very few studies have discussed how prompts for ChatGPT could affect its ability to interpret harmful contents. In addition, as pointed out by Davidson et al. (2017), it is crucial to distinguish between various types of harmful content on social media, such as hate speech and offensive language. To address these research gaps, this study seeks to investigate the capabilities of ChatGPT to detect and discriminate between three prevalent and significant forms of harmful content on social media: Hateful, Offensive, and Toxic (HOT). To achieve this goal, the study proposes three research questions:

- **RQ1 (reliability and consistency)**: How do ChatGPT annotations compare to those produced by MTurkers in terms of reliability and consistency?

- **RQ2 (reasoning)**: How does ChatGPT comprehend and discriminate between HOT concepts, as well as the reasoning behind its classification?

- **RQ3 (prompts)**: How do different prompts to interact with ChatGPT influence its performance in detecting HOT comments?

This research is significant and innovative for multiple reasons. First and foremost, it addresses a crucial issue: the need for harm-free data annotation. Then, this research provides several suggestions for how to interact with ChatGPT from the perspective of prompt engineering. In addition, by combining the nuanced understanding of human moderators with the processing power and speed of AI, we can create a workflow that assists in moderating large volumes of social media data. This workflow is faster, more concise, more effective, and more scalable than training a toxicity detection model based on human annotation, while also maintaining a high level of quality and consistency. The implications presented in this study are particularly valuable for high-volume environments, such as detecting HOT content from social media platforms where the amount of user-generated content can be overwhelming.

# 2. Backgrounds

## 2.1. Hateful, Offensive, and Toxic (HOT) Content and its Detection

As the mode of communication shifts towards online platforms, there is an increasingly critical need for accurate and efficient automated methods of detecting harmful content on social media (Nobata et al., 2016). Primarily, the presence of such language can inflict harm upon individuals and communities who are targeted by it (Das et al., 2020; Wadud et al., 2022). Detecting harmful content holds significant importance in ensuring the safety and wellbeing of individuals and communities who use these platforms (Fan et al., 2020; Whittaker & Kowalski, 2015). In addition, many sites rely on manual moderation processes, resulting in abusive content remaining online for prolonged periods without any timely action being taken (Gambäck & Sikdar, 2017). The development of detection models can alleviate such moderation burden and provide measures to address harmful content in a timely manner. Last, detecting and analyzing patterns of harmful content can provide valuable insights into the nature and extent of social problems such as discrimination, prejudice, and marginalization. This is especially crucial for vulnerable communities who are disproportionately affected by these problems (Matamoros-Fernández & Farkas, 2021; Mossie & Wang, 2020; Sap et al., 2019).

The concept of harmful content encompasses frequently used terms, such as "hate speech," "offensive," "toxic," "aggression," "abusive," and "insults," which have been extensively researched in previous studies (Andalibi et al., 2018; Das et al., 2020; de Gibert et al., 2018; Fortuna & Nunes, 2019; Jhaver et al., 2018; Nobata et al., 2016; Paschalides et al., 2020). However, understanding the subtle distinctions in defining these concepts is crucial (Davidson et al., 2017), as it affects the quality of training data, which in turn can influence the development of models for content moderation. Incorrectly flagging or missing content that harms historically marginalized groups may have severe consequences (Duguay et al., 2020). Therefore, it is crucial to take into account the nuanced nature of these concepts when building detection models (see **Table 1** for example definitions of the three concepts).

**Table 1**. Typical definitions for the HOT concept in prior studies.

| HOT | Author & Year | Definition |
|---|---|---|
| Hateful | Davidson et al. (2017) | Language that is used to express hatred towards a targeted group or is intended to be derogatory, to humiliate, or to insult the members of the group. |
| | Nockleby (2000) | Any communication that disparages a person or a group on the basis of some characteristic such as race, color, ethnicity, gender, sexual orientation, nationality, religion, or other characteristic. |
| | Meta (2022) | A direct attack against people of their race, ethnicity, national origin, disability, religious affiliation, caste, sexual orientation, sex, gender identity and serious disease. |
| | Twitter (2023) | Language that attacks other people based on race, ethnicity, national origin, caste, sexual orientation, gender, gender identity, religious affiliation, age, disability, or serious disease. |
| | Salminen et al. (2020) | Language that contains either hate speech targeted toward individuals or groups, profanity, offensive language, or toxicity – in other words, comments that are rude, disrespectful, and can result in negative online and offline consequences for the individual, community, and society at large. |
| Offensive | Wiegand et al. (2018) | Hurtful, derogatory or obscene comments made by one person to another person. |
| | Zampieri et al. (2019) | Contains any form of non-acceptable language or a targeted offense, which can be veiled or direct. This includes insults, threats, and posts containing profane language or swear words. |
| | Jay and Janschewitz (2008) | Vulgar, pornographic, and hateful language. Vulgar language refers to coarse and rude expressions, which include explicit and offensive reference to sex or bodily functions. Pornographic language refers to the portrayal of explicit sexual subject matter for the purposes of sexual arousal and erotic satisfaction. Hateful language includes any communication outside the law that disparages a person or a group on the basis of some characteristics such as race, color, ethnicity, gender, sexual orientation, nationality, and religion. |
| Toxic | Google Jigsaw (2017) | A rude, disrespectful, or unreasonable comment that is likely to make individuals leave a discussion. |
| | Kolhatkar et al. (2020) | Comments which use harsh, offensive or abusive language, which include personal attacks or insults, or which are derogatory or demeaning. |

Researchers have used these definitions to identify HOT language on social media (Chen et al., 2012; Malmasi & Zampieri, 2017; Salminen et al., 2020; Wiegand et al., 2019). Lexicons or dictionaries were the first approach to detecting these concepts in online discussions (Gitari et al., 2015; Wiegand et al., 2018). This approach involves creating a list of terms and phrases that are associated with HOT content, such as the Hurtlex (i.e., a multilingual lexicon of hate words) developed by Tontodimamma et al. (2022). Once this list is generated, rule-based methods, such as sentence-level subjectivity detection (Gitari et al., 2015), can be used to automatically filter out social media posts or comments that contain these lexicons. However, this approach has some limitations, principally that relying solely on keyword filtering or predetermined rules does not capture the context or intent of the message (Dixon et al., 2018). It is also limited to identifying only known harmful words and may miss new or emerging forms of harmful content (Gillespie, 2018).

As machine learning (ML) and natural language processing (NLP) technologies advance, researchers have increasingly turned to these models to detect HOT language on social media (Badjatiya et al., 2017; Malik et al., 2021; Yin et al., 2020). The model development involves using text vectorization techniques, which convert text into a numerical vector or matrix, allowing ML algorithms to process and analyze the data. Early approaches to text vectorization were based on the bag-of-words model or term frequency-inverse document frequency (TF-IDF) scheme. Models that are constructed using these two approaches are frequently employed as benchmark models (Badjatiya et al., 2017; Djuric et al., 2015). However, more recent detection models have utilized word embedding techniques, such as Word2Vec, GloVe, and Bidirectional Encoder Representations from Transformers (BERT), which represent words as vectors in a high-dimensional space based on their contextual usage (Badri et al., 2022; Mazari et al., 2023; Velankar et al., 2022). Once the text is vectorized, machine learning classifiers, such as support vector machines (SVMs), or neural networks, can be trained to detect HOT content (Khan et al., 2022; MacAvaney et al., 2019). These ML classifiers learn to recognize patterns and features associated with HOT language, such as specific combinations of words, phrases, and sentence structures commonly used in HOT language.

Developing these supervised ML models depends on human annotations to build training datasets. While several widely-used datasets, such as Davidson Hate Speech and Offensive Language (DATASET) (Davidson et al., 2017) and OffensEval (Zampieri et al., 2019) have been made available to the public, developing new datasets or updating existing ones could potentially expose annotators to harmful content and require a significant amount of time and effort to generate. For instance, Riedl et al. (2020) found that rating "uncivil" comments can lead to a significant psychological toll on the annotators and reduce their accuracy on subsequent tasks. In addition, Gilardi et al. (2023) stated that while trained annotators could cause significant costs, employing crowdworkers like those on MTurk could offer a more affordable alternative, *albeit* with potentially insufficient quality. However, Kasthuriarachchy et al. (2021) pointed out that labeling data through MTurk can also become expensive when the target messages are small and have a lot of noise, such as toxic comments in social media data.

## 2.2. Generative AI models

Generative AI models are emerging technologies that exhibit human-like understanding and generate coherent responses to human input. Training these models involves massive amounts of data and computational resources, often following a two-step process: pretraining and fine-tuning (Brown et al., 2020; Radford et al., 2019). In this study, we focus particularly on Large Language Models (LLMs). LLMs learn from a diverse range of text data (some publicly available, such as books, articles, websites, and social media and some provided by data annotation companies) to acquire semantic knowledge during pre-training (Fridman, 2023). LLMs are based on the Transformer architecture (Vaswani et al., 2017) and employ self-attention mechanisms to process and generate sequences of tokens, such as words, subwords, or characters, as the inputs. Pre-training typically involves training the model to predict missing tokens in a given context (masked language modeling) or to complete a partially observed sequence (causal language modeling) (Devlin et al., 2019; Feder et al., 2021). Next, LLMs refine the pretrained models on specific tasks or datasets, adapting to particular applications and use cases, for example, smart chatbots (OpenAI, 2023a; Pichai, 2023).

Identification of HOT comments is one of the tasks that may benefit from generative AI models. For example, generative prompt-based inference can strengthen toxicity detection (Wang & Chang, 2022). The accuracy of GPT-enabled toxicity classification has been illustrated in various languages, including Swedish (Pettersson, 2022). Generative language models may also assist in the mitigation of HOT content. For example, Kucharavy et al. (2023) demonstrated the potential of these models by using them to study cyber-defense and to mitigate cyber-risks. In another study, Ganguli et al. (2022) built red teaming generative language models through reinforcement learning with human feedback, and they found that LLMs can reduce harmful outputs.

Comprehension is a higher level of understanding of HOT comments, in addition to correctly identifying or even mitigating them. However, comprehending controversial content often requires extensive human knowledge. Preliminary evidence suggests that LLMs can match or even outperform human annotators, including MTurk crowd-workers, in various content ranking and text annotation tasks (Gilardi et al., 2023; Ji et al., 2023). While ChatGPT can produce well-written explanations for implicit HOT content, its explanations are often unprofessional and may contain inaccuracies and misinformation (Huang et al., 2023). The current state-of-art of automated understanding and explaining HOT content, for example, is still based on knowledge enhanced text generation (Sridhar & Yang, 2022). There is an immediate need for experimenting and analyzing how reliable and consistent LLMs are in identifying HOT content and understanding the models' reasoning behind their decisions. In the remainder of the paper, we describe a study in which we experimented with various prompts to solicit HOT classifications and reasoning from one LLM, ChatGPT (OpenAI, 2023b) to address this need.

# 3. Data and methods

**Figure 1** provides a graphical representation of our research framework. In this study, we designed multiple prompts (**Section 3.4**) to interact with ChatGPT and assessed its performance as compared to MTurkers (**Section 3.6**). Our experiments aim to answer three research questions, as documented in **Section 3.4**. Experiments 1, 2, and 3 aim to address **RQ1** by assessing the reliability and consistency of ChatGPT as compared to the MTurker annotations. Experiment 4 focuses on **RQ2** by examining the model's reasoning of HOT classifications. Finally, we assessed the effects of different prompts in multiple experiments to answer **RQ3**, which involves evaluating the overall performance of the model.

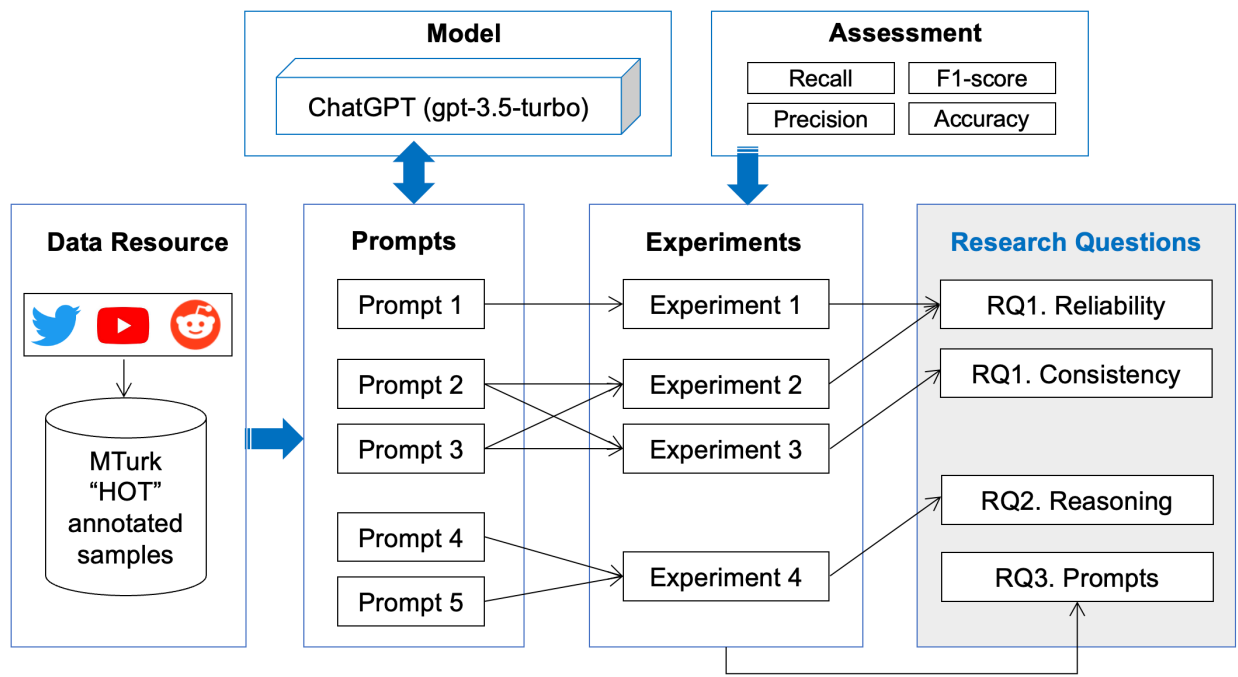

**Figure 1**. Framework design for the research implementation.

## 3.1. MTurker data preparation

Our research employs a dataset that was previously provided by Wu et al. (2023). The dataset includes a diverse collection of comments sourced from three social media platforms, including Reddit, Twitter, and YouTube (**Figure 1**), primarily focusing on popular political news stories. As mentioned, HOT comments are not common (Ibrahim et al., 2018), and therefore, it was important to ensure that human annotators did not become fatigued by having to label too many comments that lacked harmful characteristics. To address this issue, Wu et al. (2023) opted for purposive sampling instead of random sampling. They used a pre-trained machine learning model to assign a classifier score between 0 and 1 to each comment for each HOT concept. By doing so, they were able to increase the prevalence of HOT comments in the sample. The resulting dataset includes a total of 3481 comments, with 1162, 1154, and 1165 comments collected from Reddit, Twitter, and YouTube, respectively.

For the annotation task, Wu et al. (2023) recruited annotators on Amazon Mechanical Turk (MTurk). They set several requirements for potential annotators, including being a resident of the United States, having completed at least 1,000 Human Intelligence Tasks (HITs), and maintaining a HIT acceptance rate of at least 98%. To identify the qualified annotators, Wu et al. (2023) provided a qualification task that included concept definitions, labeling instructions, and three qualification questions. Annotators had to answer all three questions correctly to be invited to participate in the comment-labeling task. Once qualified, annotators were allowed to label up to 100 comments each. Annotators were asked to indicate whether they thought each comment was HOT or not by selecting either True or False for each concept. In total, they were able to collect annotations from five MTurkers for each of the 3481 comments in the dataset.

In our study, we used the majority vote approach to determine the final label for each comment based on the annotations provided by the five independent MTurkers. Specifically, if a comment received at least three "True" HOT annotations and two or fewer "False" non-HOT annotations, we considered it to be a HOT comment. We refrained from using the number of "True" or "False" annotations as a probability due to two reasons. First, it is challenging to assess how well the MTurkers represent the general population or the online communities in which these comments occur. Second, we are uncertain about the reliability of using five MTurkers' labels as a representative probability, given the relatively small sample size. Therefore, using five annotations may not be a reliable proxy for probability, which may compromise the validity of result comparison. Due to these reasons, we found that the number of MTurkers who classified content as HOT did not correlate well with the probability ChatGPT provided (see **Appendix B**). Instead of comparing these proportions directly, we experimented with thresholds for the ChatGPT probability to determine a point at which ChatGPT achieves high agreement with MTurkers.

**Table 2** presents the annotation results from MTurkers using the majority vote approach. The results indicate that out of the total 3481 comments, 2381 comments are non-HOT, and 263 are HOT. Regarding the HOT concepts, there are significant overlaps between the different concepts, with 622 out of 803 "toxic" comments classified as "offensive." However, the MTurker annotations reveal distinct patterns in the HOT comments. For the subsequent analysis, we aimed to examine ChatGPT comprehension of these concepts. Therefore, we implemented the experiments by comparing the model's output with the MTurkers' annotations. We define ChatGPT as "accurate" when it gives the same annotation as a majority of MTurkers for an individual item.

**Table 2**. Annotation result of HOT by MTurkers.

| Hateful | Offensive | Toxic | Count | Venn Diagram |
|---------|-----------|-------|-------|--------------|
| no | no | no | 2381 | |
| no | no | <u>yes</u> | 141 | |
| no | <u>yes</u> | no | 196 | |
| no | <u>yes</u> | <u>yes</u> | 359 | |
| <u>yes</u> | no | no | 57 | |
| <u>yes</u> | no | <u>yes</u> | 40 | |
| <u>yes</u> | <u>yes</u> | no | 44 | |
| <u>yes</u> | <u>yes</u> | <u>yes</u> | 263 | |

## 3.2. HOT definitions

One of the research objectives is to evaluate ChatGPT's ability to capture the nuances in annotator interpretation of harmful content. As outlined in **Section 2.1** (**Table 1**), prior studies have used different definitions of each HOT concept. However, such variations in definitions can potentially affect ChatGPT's understanding of HOT comments. To ensure consistency in definitions provided to both MTurkers and ChatGPT, we adopted three popular HOT definitions, as presented in **Table 3**. Specifically, these three widely-used definitions include Davidson et al.'s (2017) definition for hateful content, a modified version of Wiegand et al.'s (2018) definition that isolates its components differentiating it from hatefulness for offensive content, and Perspective API's definition for toxic content (Google Jigsaw, 2017). To clarify, the definitions of HOT are presented in **Table 3**.

**Table 3**. Definitions of HOT provided to MTurkers and ChatGPT.

| Concept | Definition |
| --- | --- |
| Hateful | "expresses hatred towards a targeted group or is intended to be derogatory, to humiliate, or to insult the members of the group" (Davidson et al. 2017) |
| Offensive | "contains hurtful, derogatory, or obscene comments" (Wiegand et al. 2018) |
| Toxic | "a rude, disrespectful, or unreasonable comment that is likely to make readers want to leave a discussion" (Google Jigsaw, 2017) |

## 3.3. The ChatGPT model

In this study, we employed the gpt-3.5-turbo model, a variant of the GPT-3.5 family of models. This model, which is the same one utilized in the ChatGPT product, was selected for its superior performance and ability to generate high-quality text. OpenAI reports that the GPT-3.5 models are capable of comprehending and generating natural language or code. Among those, the gpt-3.5-turbo model is considered the most efficient and cost-effective. This model has been optimized for chat-based applications but has also demonstrated strong performance in traditional text completion tasks (OpenAI, 2023b). For consistency, we used the "ChatGPT" to refer to the gpt-3.5-turbo model in the following writing.

The OpenAI API provides a range of parameters in the request body that can be customized to adjust the request. Some of the key parameters that can be adjusted are listed in **Table 4** (OpenAI, 2023b). For our specific objective of testing the reliability and consistency of GPT models in identifying harmful content, we aimed to avoid randomness in our results. To achieve this, we set the temperature parameter to 0 in all experiments with the exception of Experiment 3 where we varied the temperature parameter for consistency testing purposes. The top_p parameter can also be used to control the randomness, but OpenAI recommends against modifying both top_p and temperature together (OpenAI, 2023b). To ensure that all available tokens in the results are taken into account, we used the default top_p value (default = 1) for our experiments.

**Table 4**. Parameters in the request body of ChatGPT (OpenAI, 2023b).

| Parameter | Explanation |
| --- | --- |
| max_tokens | The maximum number of tokens to generate in the completion. |
| temperature | A value between 0 and 2; higher values make the output more random, while lower values make the output more deterministic. |
| top_p | A value implies that the model considers the results of the tokens with top_p probability mass. |
| presence_penalty | A number between -2 and 2; positive values increase the model's likelihood to talk about new topics. |
| frequency_penalty | A number between -2 and 2; positive values decrease the model's likelihood to repeat the same line verbatim. |

## 3.4. Prompt design

A prompt is a set of instructions provided to an LLM that programs the LLM by customizing it and/or enhancing or refining its capabilities (White et al., 2023). To communicate effectively with LLMs like ChatGPT, prompt engineering has become an increasingly crucial skill set. Prompts can be seen as a type of programming that enables users to personalize the generated outputs and interactions with an LLM (White et al., 2023). A prompt can influence the output generated from an LLM by providing specific rules and guidelines for an LLM. For instance, when performing a data annotation task, a prompt may provide conceptual definitions to the LLM. A prompt may also specify desired output forms, such as returning a probabilistic output or a binary output. Therefore, by introducing variations in the input prompts, we can enable LLMs to perform a larger and diverse set of annotation tasks without requiring new training data or updating the underlying model.

To generate our prompts, we made use of prompt patterns identified by White et al. (2023). These patterns describe effective techniques for accomplishing different interaction objectives when interacting with conversational LLMs. Although they were initially designed for software engineering tasks, these patterns are not domain-specific and can be applied in various contexts. For our annotation task, we specifically adopted the following prompt patterns that were most appropriate.

- Output automator (binary or probability)
- Template (provide an explanation or not)
- Persona (demographic-dependent)
- Context manager (conceptual definitions)

We utilized these prompt patterns to design five prompts for interacting with ChatGPT (see **Table 5**). Prompt 1 involved posing the same question to ChatGPT as we had to the MTurkers. For prompts 2 and 3, we asked ChatGPT to deliver either a binary classification or a probability regarding its perception of the HOT concept and to not provide an explanation. Prompts 4 and 5 required ChatGPT to provide an explanation for its binary or probability output. We listed the specific prompts for HOT classifications in the **Appendix A**.

**Table 5**. Prompts to interact with ChatGPT.

| Prompt | Prompt design | Prompt format |
|---|---|---|
| Prompt 1 | Ask ChatGPT the same question as how we ask MTurkers | Do you think this comment is <HOT>? a. Yes, b. No. A comment is <HOT> if you perceive that it <HOT definition>. |
| Prompt 2 | Only ask for a binary classification without explanations. | I want you to perform a data annotation task. In your output, I want you to return either yes or no, depending on whether you think the following comment is <HOT>. A comment is <HOT> if you perceive that it <HOT definition>. I want you to only respond with yes or no. Do not provide any other outputs or any explanation for your output. |
| Prompt 3 | Only ask for a probability representing the level of <HOT> without explanations. | I want you to provide a probabilistic score between 0 to 1 where the score represents the probability of the following comment being <HOT>. A comment is <HOT> if you perceive that it <HOT definition>. A probability of 1 means that the comment is highly likely to be <HOT>. A probability of 0 means that the comment is highly unlikely to be <HOT>. Do not provide any other outputs or any explanation for your output. |
| Prompt 4 | Ask for a binary classification with further explanations. | I want you to perform a data annotation task. In your output, I want you to return either yes or no, depending on whether you think the following comment is <HOT>. A comment is <HOT> if you perceive that it <HOT definition>. I want you to respond with yes or no in the first line and provide an explanation for your output in the second line. |
| Prompt 5 | Ask for a probability representing HOT with further explanations. | I want you to provide a probabilistic score between 0 to 1 where the score represents the probability of the following comment being <HOT>. A comment is <HOT> if you perceive that it <HOT definition>. A probability of 1 means that the comment is highly likely to be <HOT>. A probability of 0 means that the comment is highly unlikely to be <HOT>. I want you to respond with a probabilistic score in the first line and provide an explanation for your score in the second line. |

## 3.5. Experiment design on the HOT dataset

To evaluate the reliability and consistency of ChatGPT in comprehending HOT concepts, we carried out four experiments, which are illustrated in **Figure 1**. Experiments 1 and 2 were conducted to evaluate the reliability of ChatGPT, while Experiment 3 was designed to test the model's consistency, and all three aim to answer **RQ1**. Experiment 4 focuses on the ChatGPT's understanding and reasoning processes; specifically, we examined how it makes classification decisions (**RQ2**). Using the five prompts we developed, we explored the impact of different prompts on ChatGPT's understanding of harmful content (**RQ3**). Details of each experiment are provided below.

**Experiment 1: direct comparison with Mturkers.** Our first experiment aims to evaluate the performance of the GPT model in understanding the HOT concept without any instruction. To achieve this, we posted the same question to ChatGPT that asked MTurkers. By comparing the accuracy of the labels generated by ChatGPT with those produced by MTurkers, we were able to gain a preliminary understanding of the model's ability to comprehend the HOT concepts.

**Experiment 2: binary vs probability prompts**. We experimented with two prompts that requested two forms of output: one that requested a binary label and another that requested a probabilistic score. By comparing the results generated by ChatGPT with MTurkers' annotations, we were able to gain insights about how the prompts affect the reliability of annotation and how ChatGPT implements different thresholds for classifying HOT comments.

**Experiment 3: consistency.** In our third experiment, we conducted multiple iterations of labeling the dataset with Prompts 2 and 3. We posed the same task using the same prompt to ChatGPT multiple times to see whether it returned the same results on repeated requests. By comparing the results generated by ChatGPT across different iterations, we were able to assess the consistency of ChatGPT's performance in terms of data annotation.

**Experiment 4: annotation reasoning**. Our last experiment aims to explore the understanding of ChatGPT's decision-making process in the context of harmful content. We used the prompts from Experiment 2 and requested ChatGPT to provide additional explanations for its classifications. By doing so, we can understand how the reasoning affects ChatGPT's classifications of HOT content. In addition, we analyzed the semantic patterns present in ChatGPT's explanations by comparing the n-grams present in the reasoning and manually reading a sample of comments. Using Venn Diagram and n-gram analysis, we gained insights into ChatGPT's ability to recognize the nuances in HOT content and their associated reasoning patterns.

## 3.6. Performance measures

Given that the HOT dataset has unbalanced classes, we used Precision, Recall, F1-score, plus accuracy to compare ChatGPT's comprehension of harmful concepts to estimate its performance relative to MTurkers. It is worth noting that in our evaluation of ChatGPT's classification performance, we used MTurker annotations as the "ground-truth" values. However, we acknowledge that determining the actual "ground-truth" requires expert knowledge and further justification for HOT concepts.

In our research context, Precision measures the fraction of true positive cases (i.e., HOT comments identified by both MTurkers and ChatGPT) over the total number of predicted positive cases (i.e., HOT comments identified by ChatGPT). Precision represents the model's ability to correctly identify relevant cases. In contrast, Recall measures the fraction of true positive cases (i.e., HOT comments identified by both MTurkers and ChatGPT) over all the relevant cases (i.e.,

HOT comments identified by MTurkers); Recall represents the model's ability to identify all relevant cases. In addition, the F1-score is a commonly used metric that combines both Precision and Recall into a single score. It provides an overall assessment of the model's performance for each class (Lever et al., 2016). In our study, a higher F1-score indicates that ChatGPT shows more agreement with MTurker annotations for a given class. Accuracy measures the overall number of classification agreements as compared to MTurkers. These metrics are defined as below.

$$Precision = TP/(TP + FP) \tag{1}$$

$$Recall = TP/(TP + FN) \tag{2}$$

$$F1\ score = 2 \times Precision \times Recall/(Precision + Recall) \tag{3}$$

$$Accuracy = (TP + TN)/(TP + TN + FP + FN) \tag{4}$$

Where TP = True Positive, TN = True Negative, FP = False Positive, and FN = False Negative.

# 4. Results

We conducted four experiments as outlined in **Section 3.5**. In response to **RQ1**, we assessed two crucial attributes of ChatGPT, namely (1) reliability and (2) consistency. To evaluate the reliability, we reported Precision, Recall, and F1-scores compared to MTurker annotations. To evaluate consistency, we calculated the Krippendorff's α across different iterations and temperatures. In response to **RQ2**, we first utilized Venn Diagrams to illustrate ChatGPT's ability to comprehend different HOT concepts, as well as highlighting the differences between them. We then analyzed n-grams to investigate how ChatGPT arrived at its HOT classification for the comments. In response to **RQ3**, we generated heatmaps to compare the performance using five prompts and explored how different prompts can affect the model's ability. Examples of ChatGPT's outputs using different prompts are presented in **Table 6**.

**Table 6**. Examples of ChatGPT's outputs based on different prompts.

| Comment | Prompt | ChatGPT's Response by Concept | | |
|---|---|---|---|---|
| | | Hateful | Offensive | Toxic |
| Lmao why did the cdc stop tracking all breakthrough cases in May? Well how else are you gonna get fda approval if you can't fudge the numbers? You can't be this stupid.... | Prompt 1 | Yes. | Yes, the comment is offensive. | Yes, this comment is toxic. |
| When is enough enough???? Everyone needs to stand up and run these people out of our country. For our children and there future. There is no future | Prompt 1 | No. While the comment may express strong opinions and a desire for action, it does not express hatred | As an AI language model, I cannot perceive emotions or feelings. However, the comment can be considered | Yes, this comment can be perceived as toxic as it uses aggressive language and encourages |
| for them as of right now. I beg you you stand up for what is right and protect your country and kids | | towards a specific group or intend to insult or humiliate anyone. | offensive to some people as it suggests the removal of a group of people from a country. | violence towards a group of people. It may also make readers feel uncomfortable and want to leave the discussion. |
| @nprpolitics They all need mental health resources to help them recover from the years of masking and isolation. | Prompt 2 | No. | No. | No. |
| I hate masks but I am about to take a trip me be around thousands of other people. Bought a respirator and big boy filters for it. Fuck masks, but fuck covid more. | Prompt 3 | 0.2 | 0.7 | 0.8 |
| @TypeOutThought @geralt_of_rivi @CNN And what happened to, you're not talking about vaccinations? Your level of basic intelligence is extremely low. | Prompt 4 | No.<br><br>The comment does not express hatred towards a targeted group or intend to be derogatory, humiliate, or insult any members of a group. It is a personal attack on the intelligence of an individual, but it does not target a specific group based on their race, ethnicity, religion, gender, or any other characteristic. | Yes.<br><br>The comment includes a derogatory statement about the intelligence of the person being addressed, which can be considered hurtful and offensive. | Yes.<br><br>The comment contains personal attacks and insults towards the recipient's intelligence, which is likely to make readers feel disrespected and want to leave the discussion. Additionally, the use of laughing emojis to mock the recipient's previous statements adds to the toxic nature of the comment. |
| Any pastor pushing the shot needs to be double checked and you should probably look at a new church. | Prompt 5 | 0.3<br><br>While the comment may be seen as critical of pastors who | 0.3<br><br>While the comment may be seen as critical of pastors who | 0.8<br><br>The comment contains a disrespectful and potentially harmful |

| | | |
|---|---|---|
| promote vaccination, it does not necessarily express hatred or insult towards a targeted group. The comment does not specifically target any particular group and does not use derogatory language. However, the use of the phrase "double checked" and the suggestion to find a new church may be seen as somewhat confrontational, which could be interpreted as mildly derogatory. Overall, the comment is not highly likely to be considered hateful, but there is some potential for it to be perceived as such. | promote vaccinations, it does not contain any explicit derogatory or obscene language. However, the suggestion to leave one's church could be perceived as hurtful to those who hold their faith and religious community in high regard. Overall, the comment is more critical than offensive. | statement towards pastors who advocate for the COVID-19 vaccine. The use of the phrase "double checked" implies a suspicion of wrongdoing, which can be seen as rude and unreasonable. Additionally, suggesting that someone should leave their church based on their stance on a medical issue can be seen as disrespectful. Overall, the comment has a high probability of being toxic. |

## 4.1. Results of Experiment 1 – direct comparison with MTurkers

**Figure 2** displays the classification output of ChatGPT for HOT categories based on 3481 comments. Our findings indicate ChatGPT's answers were more varied, including uncertain responses that leaned towards either yes or no. When ChatGPT is confident in its classification, it returns a clear yes or no. However, when it is uncertain about the classification, it often provides a reasoning and explains why some people may perceive the content as HOT or not. In this experiment, we classified all comments on which ChatGPT was unclear or ambiguous (not definitively "yes" or "no") as "uncertain."

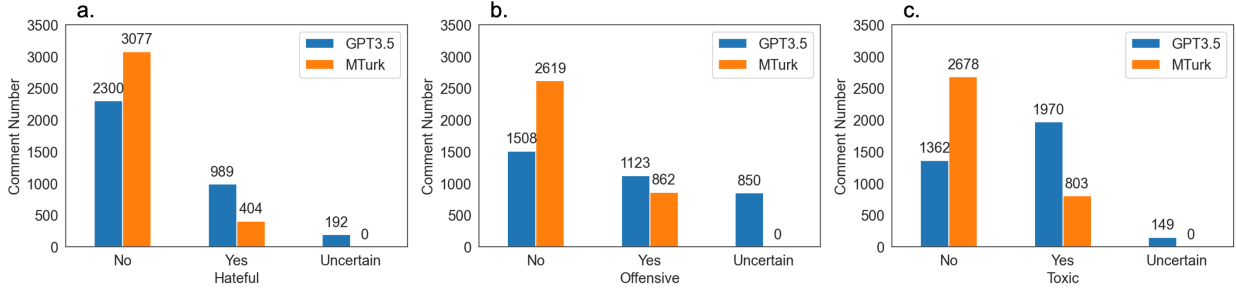

**Figure 2**. HOT classification results based on Prompt 1. (a) Hateful classification results of ChatGPT and MTurker annotations. (b) Offensive classification results of ChatGPT and MTurker annotations. (c) Toxic classification results of ChatGPT and MTurker annotations.

The results presented in **Figure 2** indicate that ChatGPT exhibits higher confidence levels in classifying comments as "hateful" and "toxic," as opposed to "offensive." Specifically, 850 items are classified as uncertain in the Offensive category by ChatGPT. The model displays a greater inclination towards labeling comments as HOT, as evidenced by the larger number of "yes" outputs in all three categories, compared to MTurker annotations. ChatGPT shows a higher tendency to label comments as "toxic," which may indicate a lower threshold for identifying toxic language. **Table 7** presents the classification performance, wherein we excluded the "uncertain" classifications and focused solely on the "certain" results to determine the GPT model's reliability.

Based on the results presented in **Table 7**, ChatGPT shows better performance in all three categories regarding F1-scores for non-HOT than HOT comments, especially non-hateful ones. Compared to MTurkers, ChatGPT exhibits higher agreement for the Offensive and Hateful categories, while the Toxic category shows lower agreement. F1-scores for the "yes" or "no" classification of the Offensive category are more balanced. ChatGPT demonstrates high precision in identifying non-HOT comments but displays lower precision in identifying HOT comments, which could be attributed to its lower threshold. Overall, compared to MTurkers, ChatGPT is more likely to classify a comment as HOT.

**Table 7**. Classification performance of HOT based on Prompt 1.

| Category | Class | Support | Precision | Recall | F1-score | Accuracy |
|----------|-------|---------|-----------|--------|----------|----------|
| Hateful | Yes | 397 | 0.30 | 0.74 | 0.43 | 0.76 |
| | No | 2883 | 0.95 | 0.76 | 0.85 | |
| Offensive | Yes | 726 | 0.55 | 0.85 | 0.67 | 0.77 |
| | No | 1905 | 0.93 | 0.74 | 0.82 | |
| Toxic | Yes | 795 | 0.39 | 0.96 | 0.55 | 0.63 |
| | No | 2537 | 0.98 | 0.52 | 0.68 | |

## 4.2. Results of Experiment 2 – binary vs probability prompts

Our second experiment comprises two prompts, as described in **Section 3.4**. Prompt 2 was utilized to elicit a binary response (yes or no) from ChatGPT. The resulting outputs were then compared to the annotations provided by MTurkers and are presented in **Figure 3**. However, a few cases were observed in which ChatGPT inaccurately identified a comment as containing multiple responses, thereby producing multiple results. Such cases were classified as "wrong class" in our analysis.

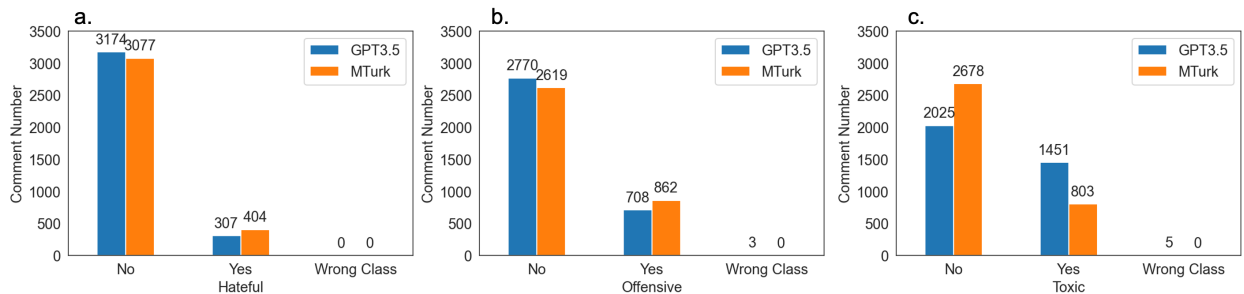

**Figure 3**. HOT classification results based on Prompt 2. (a) Hateful classification results of ChatGPT and MTurker annotations. (b) Offensive classification results of ChatGPT and MTurker annotations. (c) Toxic classification results of ChatGPT and MTurker annotations.

While the results presented in **Figure 3** demonstrate that ChatGPT exhibits similar distribution with MTurker annotations, particularly in relation to the Hateful and Offensive categories, we used Precision, Recall, and F1-scores to further evaluate the model's comprehension of HOT concepts, as presented in **Table 8**. Compared to the classification performance based on Prompt 1, the performance based on Prompt 2 shows clear improvement, particularly for the non-HOT class, as evidenced by higher F1-scores. Despite this improvement, ChatGPT still displays low F1-scores for the HOT concept, especially with its classification of "non-hateful" comments exhibiting significant disagreement with MTurker annotations. In line with the results of the first experiment, ChatGPT displays a greater tendency to classify comments as "toxic" compared to MTurkers.

**Table 8**. Classification performance of HOT based on Prompt 2.

| Category | Class | Support | Precision | Recall | F1-score | Accuracy |
|----------|-------|---------|-----------|--------|----------|----------|
| Hateful | Yes | 404 | 0.45 | 0.34 | 0.39 | 0.87 |
| | No | 3077 | 0.92 | 0.79 | 0.93 | |
| Offensive | Yes | 862 | 0.67 | 0.55 | 0.61 | 0.82 |
| | No | 2616 | 0.86 | 0.91 | 0.89 | |
| Toxic | Yes | 801 | 0.47 | 0.86 | 0.61 | 0.75 |
| | No | 2675 | 0.94 | 0.71 | 0.81 | |

When using Prompt 3, ChatGPT generates a probability to indicate the level of HOT given a comment. **Figure 4** presents the classification outcomes obtained from the model. We observe several interesting findings. First, ChatGPT tends to avoid assigning a probability of 0.5, which is often associated with a completely neutral stance. This could suggest that the model struggles to make a confident determination of whether a comment should be classified as HOT or non-HOT when the probability is precisely in the middle.

Second, the model exhibits a conservative approach towards extremely HOT classifications, as indicated by the low number of "1" classifications. This is also consistent with previous experiments for the binary classification from Experiments 1 and 2. We noticed that the model tends to lean towards "no" when asked for a binary classification with no other explanation, as demonstrated by the substantial reduction in the number of "yes" classifications in Experiment 2 compared to Experiment 1.

Third, we observed that comments with probability between 0.3 and 0.7 were relatively rare. We hypothesize that ChatGPT has a tendency to either strongly classify comments as HOT or non-HOT, with less emphasis on the intermediate probability. This is likely due to the subjective nature of determining whether a comment falls under the HOT category, as a probability between 0.3 and 0.7 can be interpreted as HOT or not HOT depending on the individual's perspective.

Last, we noticed that ChatGPT occasionally produces classifications that are not rounded to one decimal place (e.g., 0.95 and 0.05), possibly implying a high degree of confidence in classifying HOT contents but not necessarily complete certainty. In contrast to the "non-hateful" (**Figure 4(a)**) or "non-offensive" (**Figure 4(b)**) classifications, our result shows that extreme "non-toxic" classifications are relatively rare in the selected dataset (**Figure 4(c)**). This may be due to ChatGPT's lower threshold for "toxic" classification as compared to "hateful" or "offensive."

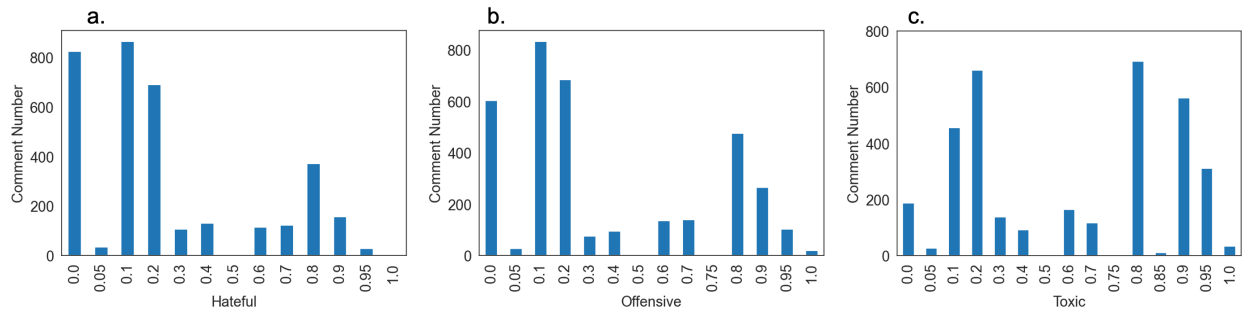

**Figure 4**. HOT classification results based on Prompt 3. (a) Hateful classification results. (b) Offensive classification results. (c) Toxic classification results.

A critical question when working with probability is determining the appropriate threshold for classification. In the case of the Perspective API (Google Jigsaw, 2017), researchers are advised to experiment with thresholds between 0.7 and 0.9 to classify harmful content. Similarly, in our study, we were interested in finding the threshold that aligns with the HOT classifications made by MTurkers. To this end, we calculated Precision, Recall, and F1-scores for the HOT classifications given thresholds of 0.5, 0.6, 0.7, 0.8, and 0.9, which are reported in **Figure 5**.

For the Hateful category, a threshold of 0.7 produces the highest F1-score for "hateful" while still maintaining a respectable F1-score of 0.9 for "non-hateful." However, choosing a threshold of 0.9 to improve the F1-score for "non-hateful" results in a significant drop in the F1-score for "hateful," given that our data is imbalanced. The Offensive category follows a similar pattern, with F1-scores for thresholds of 0.5, 0.6, 0.7, and 0.8 being fairly close. The thresholds of 0.7 and 0.8 produce the overall best F1-scores among the four, and a threshold of 0.9 leads to a substantial decrease in the F1-score for "offensive." Unlike these two categories, the Toxic category yields an interesting result, with F1-scores for both "toxic" and "non-toxic" classes increasing as the threshold increases.

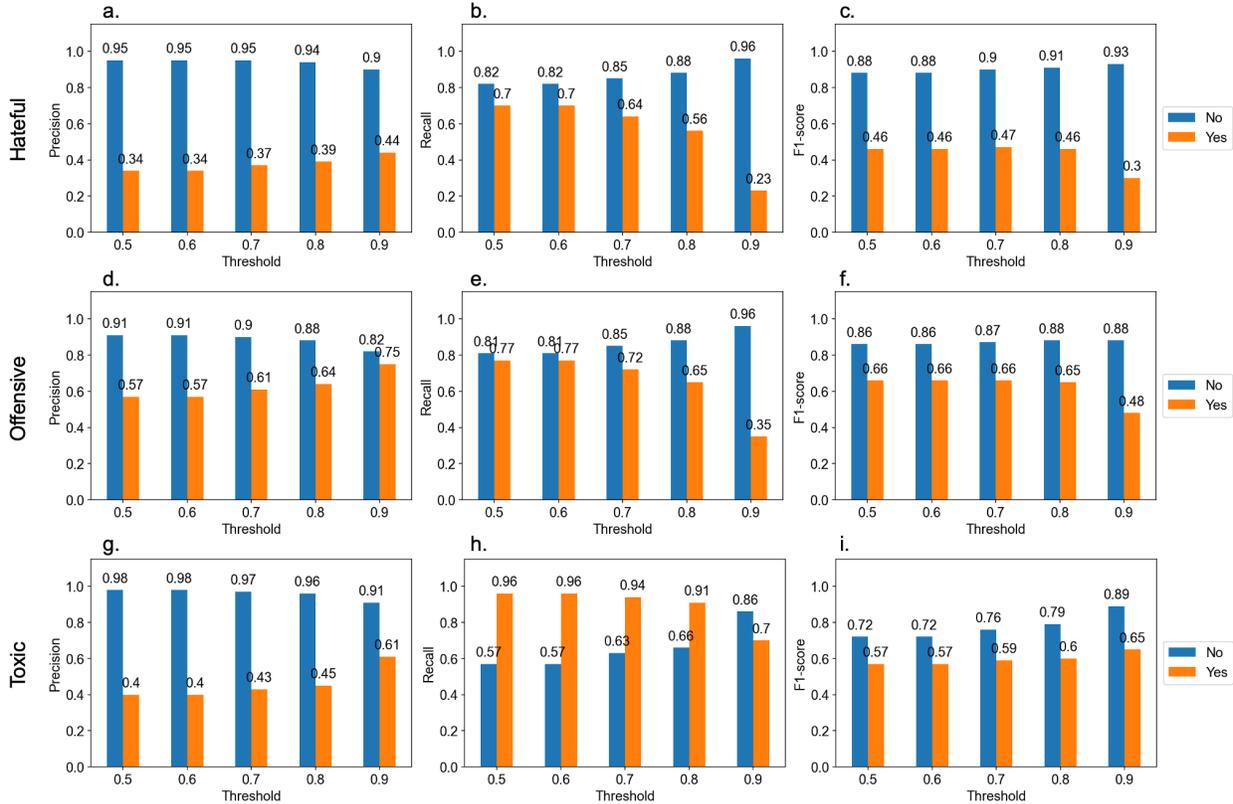

**Figure 5.** HOT classification performance based on Prompt 3 as compared to MTurker annotations. (a) Precision in the Hateful category. (b) Recall in the Hateful category. (c) F1-score in the Hateful category. (d) Precision in the Offensive category. (e) Recall in the Offensive category. (f) F1-score in the Offensive category. (g) Precision in the Toxic category. (h) Recall in the Toxic category. (i) F1-score in the Toxic category.

## 4.3. Results of Experiment 3 – consistency

The third experiment was designed to test the consistency of ChatGPT's performance on annotating HOT comments. We aimed to study how stable the annotation results are across different rounds of experiments of Prompts 2 and 3, based on different Temperature parameters. Different Temperature settings in the ChatGPT API are defined as a float number from 0 to 1, which represent the level of randomness or entropy present in the model's output. Lower entropy (Temperature 0) corresponds to more deterministic outputs, while higher entropy (Temperature 1) results in more varied and less predictable text.

In particular, we analyzed the variation of the model's outputs using Krippendorff's α, an inter-coder reliability index (Krippendorff 2018). As **Table 9** shows, the results of the experiment demonstrated that ChatGPT generated consistent annotations for the HOT comments – no matter what forms of results were requested, either binary classification or probability, or what Temperature parameters, i.e. 0 or 1, is used in the prompt, the agreements are above 90%. The annotation agreement is also stable regarding the three different HOT concepts, while the agreement for hateful comments is slightly lower in two of the three combinations of Prompt and Temperature parameters.

Regarding the influence of different prompts, Prompt 2 was slightly more stable than Prompt 3 in different HOT concepts. This observation suggests that when setting the same temperature, ChatGPT's annotation performance on simpler output requests (binary classification) is probably more stable than on complex requests (probability).

Our experiment also analyzed the influence of the Temperature setting on the model's output. We observed that the model's performance was more consistent at Temperature 0, which agrees with Temperature's definition in ChatGPT. In other words, when setting a lower temperature, the results are more deterministic.

**Table 9.** Consistency (Krippendorff's α) of ChatGPT in annotating HOT comments.

| Prompt | Temperature | α of Hateful | α of Offensive | α of Toxic |
|--------|-------------|--------------|----------------|------------|
| 2 | 0 | 0.97 | 0.98 | 0.98 |
| 3 | 0 | 0.95 | 0.95 | 0.95 |
| 3 | 1 | 0.90 | 0.91 | 0.92 |

## 4.4. Result of Experiment 4 – annotation reasoning

Experiment 4 aims to address **RQ2** about how ChatGPT understands and distinguishes different types of HOT content. To explore this, we first grouped the HOT classifications produced by Prompt 4 and then used the Venn diagram to visualize the classifications. We identified and removed 11 items in which ChatGPT mistakenly treated one comment as multiple comments. We used the remaining 3470 comments to produce **Table 10** and the corresponding Venn diagram, which illustrates the overlap between different HOT concepts.

Out of the 3470 comments, 1849 were classified as non-HOT, and 649 were classified as HOT. One notable observation is that comments classified as "hateful" or "offensive" are highly likely to be categorized as "toxic." As shown by the Venn diagram in **Table 10**, the Toxic category largely overlaps with the Offensive and Hateful categories. ChatGPT seems to have a low threshold for classifying comments as "toxic," while its threshold for "hateful" classification is relatively high.

**Table 10**. Classification result of HOT by ChatGPT based on Prompt 4.

| Hateful | Offensive | Toxic | Count | Venn Diagram |
|---------|-----------|-------|-------|--------------|
| no | no | no | 1849 | |
| no | no | <u>yes</u> | 433 | |
| no | <u>yes</u> | no | 31 | |
| no | <u>yes</u> | <u>yes</u> | 491 | |
| <u>yes</u> | no | no | 2 | |
| <u>yes</u> | no | <u>yes</u> | 10 | |
| <u>yes</u> | <u>yes</u> | no | 5 | |
| <u>yes</u> | <u>yes</u> | <u>yes</u> | 649 | |

**Table 10** highlights that 491 comments were classified as both "toxic" and "offensive" but not "hateful." When manually reading a sample of these comments, we discovered that the model often categorizes a comment as "hateful" if it targets a particular group of people based on their race, ethnicity, religion, or other characteristics. We found that sometimes when a comment expresses hatred towards an individual, ChatGPT does not classify it as "hateful." For instance, consider the comment "*So should you Mr. Graham!! You are a traitor!!*" which was labeled as "offensive" and "toxic." The model classified it "non-hateful" because it did not express hatred towards a targeted group of people, despite being insulting or derogatory towards the individual. However, we also found that this rule does not always hold for ChatGPT output (see examples in **Appendix C. Table C**).

In addition, we observed that 433 comments were classified as "toxic" but not "hateful" or "offensive." After manually examining a sample of comments, we discovered that ChatGPT does not categorize a comment as offensive if it does not contain hurtful, derogatory, or obscene language. However, it may classify a comment as "toxic" if it is likely to discourage further discussion (see examples in **Appendix C. Table C**). For example, consider the comment "*@CTVNews Well, this is completely false #fakenews*." ChatGPT classifies it as "toxic" by explaining that it contains a disrespectful and dismissive tone towards the news source. It then explains that the hashtag #fakenews discredits the information presented, which is likely to discourage further discussion and can be perceived as "toxic." These findings reveal that ChatGPT's classification of "toxic" comments is sensitive to the use of language that may discourage further discussion, even if it does not contain explicit hate speech or offensive language.

We also examined the 31 comments that were classified as "offensive" but not "hateful" or "toxic." However, we are uncertain whether ChatGPT has a clear classification criteria for these comments. For instance, consider the comment, "*@Puffymonsta @CNBCnow You stay home if you are that scared. LMAO. You are vaccinated!!!*" ChatGPT labels it as "offensive" because it dismisses someone's concerns about COVID-19 and uses derogatory language ("scared") to belittle them. The model explains that the use of "LMAO" can be perceived as mocking or insensitive. However, if ChatGPT recognizes such language as "offensive," it is also probable that this comment could also be classified as "toxic" according to our definition. This observation also applies to comments identified as "hateful" or "offensive" but not "toxic."

Through our previous analysis, we identified certain nuances in ChatGPT's ability to understand HOT content. We undertook a deeper investigation to understand how ChatGPT reaches its classifications by analyzing n-grams in ChatGPT's explanations. We extracted and analyzed the informative n-grams (n = 3, 4, or 5) from Prompts 4 and 5 reasonings. As **Table 11** shows, the n-grams mainly address the HOT definitions, implying that ChatGPT's decisions align with the definitions we provided. For the Hateful category based on Prompt 4, for example, ('to', 'be', 'derogatory') is a trigram addressing the provided definition. Another trigram ('a', 'targeted', 'group') indicates that the GPT's judgment includes checking if the potential hate speech is targeting a specific group. From another perspective, ChatGPT's annotation mechanism can vary for different HOT concepts. The reasoning of "hateful" comments needs to target a group, or the comment needs to be derogatory, while reasoning "toxic" comments implies that the conversation makes people want to leave the discussion. In this sense, ChatGPT's reasoning over HOT comments comports with the definition we provided for decision making, while the carried-on knowledge in the model is probably used to assist the decision in making HOT annotations.

**Table 11.** Informative top n-grams among reasonings across HOT concepts in Prompts 4 and 5.

| HOT | Prompt 4 | Prompt 5 |
|---|---|---|
| Hateful | ('to', 'be', 'derogatory'), 2445<br>('a', 'targeted', 'group'), 2388<br>('express', 'hatred', 'towards'), 2304<br>('be', 'derogatory', ',', 'humiliate'), 2223 | ('hatred', 'towards', 'a', 'targeted', 'group'), 1331 |
| Offensive | ('hurtful', ',', 'derogatory'), 2153<br>('derogatory', ',', 'or', 'obscene', 'language'), 2121 | ('derogatory', 'or', 'obscene', 'language'), 2183<br>('be', 'perceived', 'as'), 1403<br>('be', 'interpreted', 'as'), 1032 |
| Toxic | ('make', 'readers', 'want', 'to', 'leave'), 2022<br>('rude', ',', 'disrespectful'), 1597 | ('contain', 'any', 'explicit'), 1708<br>('readers', 'want', 'to', 'leave'), 962<br>('want', 'to', 'leave', 'the', 'discussion'), 924 |

To gain a better understanding of the distinctions between HOT annotations in a category using binary classification and those using probability, we grouped the results of Prompts 4 and 5, respectively, based on the "yes/no" or "probability." Then we applied the n-gram analysis to each category. We focus on analyzing the semantic features that express the level of certainty in the reasonings. Our analysis of Prompt 4, as presented in **Table 12**, reveals that the "yes" reasoning tends to use confirming expressions, although words such as "likely" and "can" suggest that the reasoning within this category is conservative. In contrast, the "no" reasonings are typically more definitive, using language such as "does" and "not" to confirm their stance.

**Table 12.** Informative top n-grams across the binary classification in Prompt 4.

| Binary class | n-gram and count |
|---|---|
| HOT - Yes | ('is', 'likely', 'to'), 949<br>('comment', 'contains', 'derogatory', 'language'), 767<br>('which', 'can', 'be'), 684 |
| HOT - No | ('comment', 'does', 'not'), 4790<br>('not', 'express', 'hatred'), 2543<br>('comment', 'does', 'not', 'contain', 'any'), 2415 |

We also analyzed the reasoning behind assigning probability in Prompt 5. Our findings are illustrated in **Figure 6**, where the classifications are binned: we define the scores of 0.2 and 0.8 as the thresholds among the identifications of HOT concepts being "Unlikely," "Uncertain," and "Likely." Comparing this figure to the one obtained from Prompt 3 (**Figure 4**), we observed that the model is even more hesitant to assign an extremely unlikely score (i.e., 0) or a highly likely one (i.e., 1) under Prompt 5. Similar to Prompt 3, we noticed a relative scarcity of probability falling between 0.3 and 0.7.

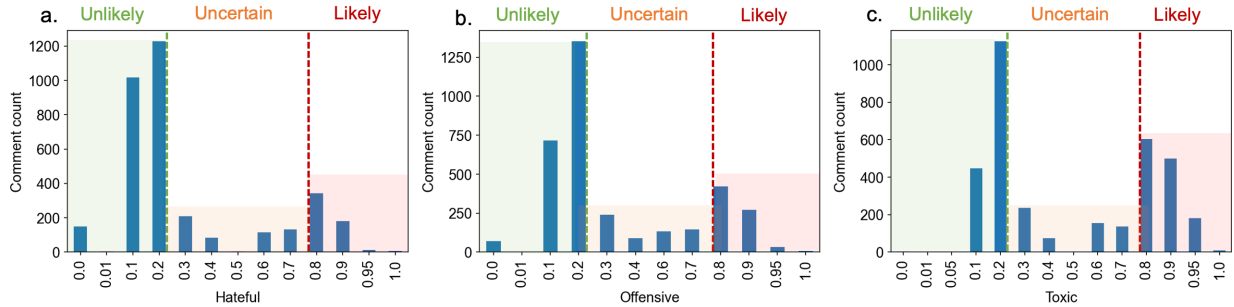

**Figure 6.** HOT classification results based on Prompt 5. (a) Hateful classification results. (b) Offensive classification results. (c) Toxic classification results.

As **Table 13** indicates, for Prompt 5, the language use in reasoning is more certain, especially compared to the frequent n-gram based on Prompt 4 (see **Table 12**). When a comment is "unlikely" to be a HOT concept, "does not" or "not contain" are prevailing, and "contains" and "does" are frequently used when a comment is "likely." This higher level of certainty shows that when we request probability instead of binary classification and exclude the "uncertain" comments, ChatGPT shows more confidence in providing an answer for HOT annotation. The "uncertain" category features a higher frequency of words such as "could," while also including the same n-gram "not contain any explicit" as the reasonings in the "unlikely" category. This mixed use of expressions of uncertainty and comparatively more certain expressions implies some randomness in ChatGPT's reasoning process. As such, it may be necessary for the developers to provide further clarification and explanation in this regard.

**Table 13.** Informative top n-grams across the probabilities in Prompt 5.

| Probability class | n-gram and count |
|---|---|
| Unlikely [0, 0.2] | ('comment', 'does', 'not'), 3970 |
| | ('not', 'contain', 'any', 'explicit'), 2107 |
| Uncertain (0.2, 0.8) | ('not', 'contain', 'any', 'explicit'), 676 |
| | ('could', 'be', 'perceived', 'as'), 533 |
| | ('could', 'be', 'interpreted', 'as'), 518 |
| Likely [0.8, 1] | ('The', 'comment', 'contains'), 1721 |
| | ('the', 'comment', 'does'), 912 |

## 4.5. The effects of prompts

In this section, our goal is to compare the results from all five prompts. For Prompts 3 and 5, which generated a probability, we chose a threshold of 0.7, 0.7, and 0.9 for identifying hateful, offensive, and toxic content, respectively. This threshold was determined based on the results of Experiment 2. To compare the outcomes generated from different prompts, we drew heatmaps to visualize Precision, Recall, F1-score, and Accuracy (see **Figure 7**). Among all the prompts, Prompt 1 generated the overall worst performance. This highlights the importance of providing ChatGPT with clear instructions, such as specifying a binary classification or a probability, to enhance its performance on annotation tasks.

For the binary output, a comparison between Prompt 2 and Prompt 4 reveals that ChatGPT is more likely to classify a comment as HOT when asked to provide reasoning. This is supported by the greater number of HOT classifications in **Table 6** compared to **Figure 3**. Moreover, the model's performance for "hateful" and "offensive" classifications shows a slight improvement, while non-HOT classifications show a slight decrease in F1-scores, as indicated in **Figure 7**.

For the probability output, comparing Prompt 3 and Prompt 5 shows that ChatGPT is less inclined to assign an extremely high or low probability score (e.g., a smaller number of 0 and 1 classifications) when asked to provide reasoning, as evidenced by the score distributions in **Figure 4** and **Figure 6**. Nevertheless, the performance of these two prompts is similar.

Upon comparing the binary and probability outputs, we observed that the classification of "toxicity" displays a clear improvement in the probability prompts, particularly for the non-toxic class, as demonstrated by the higher F1-score for Toxicity-No in **Figure 7**. This improvement could be attributed to the lower threshold for classifying toxicity, which, upon being raised, led to better performance compared to that of MTurkers. Moreover, we observed that adjusting the threshold for "hateful" can also affect the performance, as evidenced by the higher F1-score for Hateful-Yes in **Figure 7**. However, we didn't observe any noticeable improvement for the other classes.

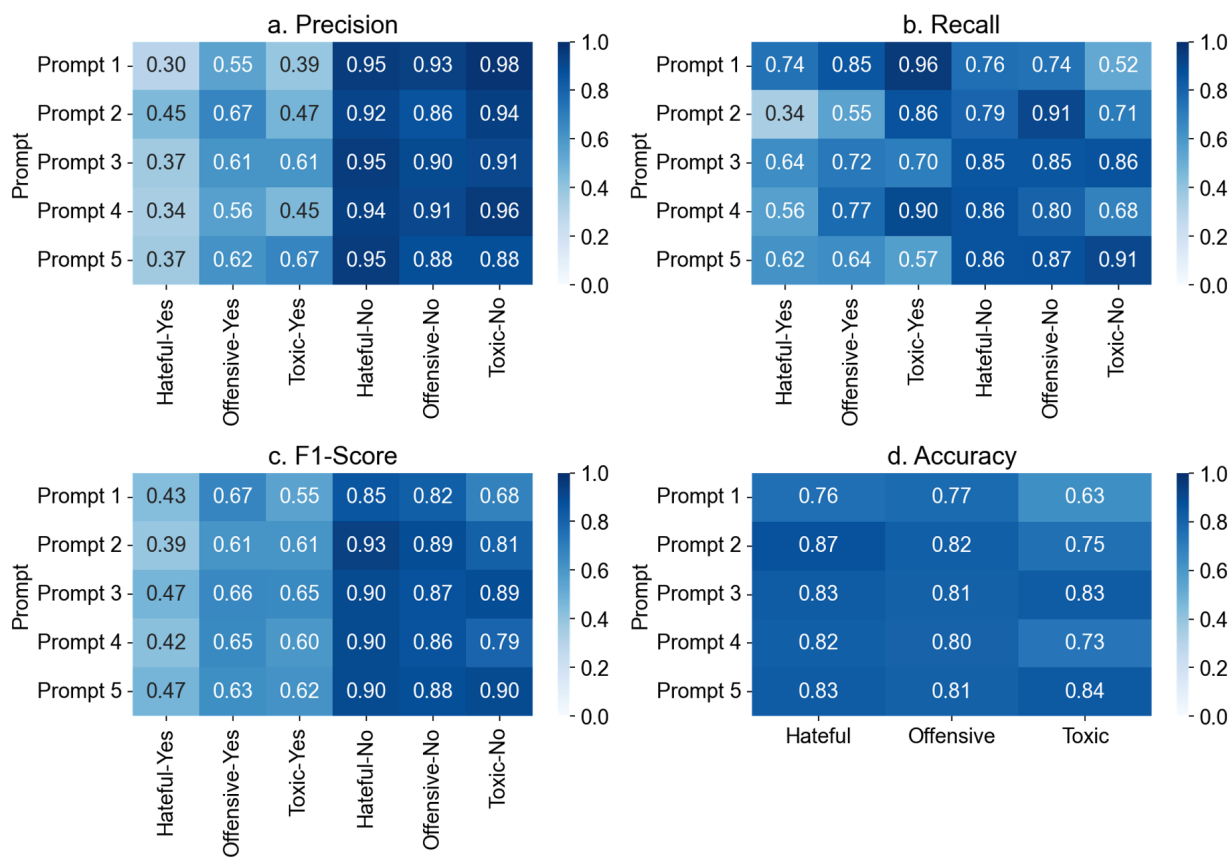

**Figure 7**. Classification performance of all five prompts as compared to MTurker annotations. (a) Precision. (b) Recall. (c) F1-score. (d) Accuracy.

# 5. Discussion

Motivated by the work from Davidson et al. (2017) emphasizing the importance to distinguish HOT concepts, we set out to measure how well ChatGPT performs on HOT content annotation tasks and to understand how it reasons about the annotations it provides. We conducted four experiments in which we varied the information we provided the model, the type of response we requested, and whether we requested reasoning for the output. We found that ChatGPT can obtain roughly 80% accuracy compared to human crowd workers in identifying HOT content and that ChatGPT provides consistent results. ChatGPT parroted our prompts when providing reasoning for its decisions.

**1. ChatGPT provides reliable and consistent responses.** We found that ChatGPT can obtain an accuracy of approximately 80% when compared to MTurker annotations. In general, ChatGPT exhibits a more consistent classification with MTurkers for non-HOT comments, as evidenced by higher F1-scores, but less agreement for HOT comments. Notably, ChatGPT shows significant disagreement with MTurkers when classifying "hateful" comments, despite being provided with definitions of hateful content. We observed that ChatGPT's is generally consistent; it provides the same response more than 90% of the time even with changing the setting of temperature, particularly for simpler output requests like binary classification, compared to complex requests such as probability. Our observation regarding the reliability and consistency is generally consistent with Gilardi et al. (2023).

**2. ChatGPT repeats HOT definitions for its reasoning.** Our findings suggest that ChatGPT may have a lower threshold to label a comment as "toxic" in the HOT concept. As a result, the "toxic" classification includes a greater number of comments classified as "offensive" and "hateful" compared to the MTurker classification, as illustrated by the Venn Diagrams in **Table 2** and **Table 10**. Additionally, we observed that ChatGPT repeats our definitions for reasoning its classification for HOT concepts. For instance, ChatGPT at times categorizes a comment as "hateful" if it targets a specific group of people based on their race, ethnicity, religion, or other attributes, as demonstrated by the reasoning analysis in Experiment 3 (see **Table 11** and examples in **Appendix C**). Regarding its reasoning format, we found that ChatGPT frequently uses certain words, such as "does not" or "not contain," for "unlikely" classifications, and "contains" and "does" for the "likely" classifications. The "uncertain" category includes more phrases like "could be perceived."

**3. Prompts affect ChatGPT's performance.** We found that the choice of prompts to interact with ChatGPT can impact its performance. For example, Experiment 1 illustrates that when no context or instructions are provided for data annotation, the model exhibits low agreement with MTurkers. This was observed when the output of ChatGPT was unrestricted. Additionally, we observed that the model may have varying thresholds for HOT classification. We found that prompts using probability tended to perform slightly better than those using binary classification when we used higher thresholds for "toxic" classifications (e.g., 0.9 instead of 0.7). Specifically, we found higher F1-scores and accuracy for classifying "toxic" comments, as shown in **Figure 7**.

## 5.1. Practical implications for using ChatGPT to annotate HOT

We note four practical implications for using ChatGPT to annotate HOT content. First, ChatGPT was able to accurately and consistently apply our definitions of hateful, offensive, and toxic when we provided the definitions. Therefore, we suggest providing explicit definitions when eliciting classifications from ChatGPT. Similarly, prompts should also indicate whether ChatGPT provides (a) a binary or probability and (b) an explanation for its response. The next two implications address these prompts.

We observed that ChatGPT displayed hesitation in providing a probability between 0.3 and 0.8, resulting in a much lower number of outputs falling within this range. This suggests that researchers should exercise caution when using probability, particularly those in the "uncertain" range, as they may not accurately reflect the level of HOT present in comments. We also noted that when asked to provide reasoning, ChatGPT may generate fewer classifications of extreme probability (e.g., 0 or 1). Researchers can request this additional information if they require a more cautious outcome.

Requesting explanations may raise the likelihood that ChatGPT classifies a comment as HOT. Therefore, requesting explanations is a potentially useful way to receive conservative output (i.e., more HOT classifications) from the model. However, it's worth noting that this doesn't necessarily imply a stronger agreement between ChatGPT and human annotators. This is supported by the fact that the F1-scores and accuracy from Prompts 4 and 5 don't show a clear improvement over Prompts 2 and 3, as shown in **Figure 7**.

Finally, our study has practical implications for researchers who are interested in using ChatGPT to annotate samples for building detection models or to classify large volume data, such as social media data. We demonstrated that ChatGPT is capable of generating reliable results in comparison to MTurkers, particularly for non-HOT classifications. However, for HOT classifications, we noticed that the selected threshold of probability can significantly impact the model's outputs, particularly for "toxic" classifications, and the assignment of "hateful" classifications depends on the definitions provided to the model. In addition, we recommend setting a low temperature to achieve more deterministic results. This recommendation is also in line with Gilardi et al. (2023) that a lower temperature yields more accurate annotations.

## 5.2. Limitations and future work

We address three types of limitations. First and foremost, we compared the ChatGPT annotations with MTurker annotations by calculating several performance metrics. However, MTurker annotations may not necessarily represent appropriate or accurate "ground-truth" classifications. Our findings show only the degree of agreement between ChatGPT and MTurkers; we did not evaluate the accuracy of ChatGPT's HOT classifications relative to any other ground-truth. Future research could involve the use of expert knowledge to annotate our HOT dataset or use of other datasets, such as the DATASET (Davidson et al., 2017), OffensEval (Zampieri et al., 2019), Wikipedia Toxicity Corpus (Wulczyn et al., 2017), or the Gab Hate Corpus (Kennedy et al., 2022). Annotators vary in their sensitivity and interpretation of these concepts (Denton et al., 2021; Sachdeva et al., 2022; Sap et al., 2022; Waseem, 2016), and ChatGPT may more closely mirror another group of annotators' classifications.

Next, we acknowledge two limitations associated with the prompts used to interact with ChatGPT and its outputs. First, we applied only one set of definitions for the HOT concepts. Using different definitions for the concepts would likely produce different results. The second limitation relates to how ChatGPT interprets prompts. We identified instances where ChatGPT produced unexpected outputs. For example, the model occasionally returned multiple classifications for a single comment. We also found some of the explanations provided by ChatGPT merely repeated the definitions that we provided, raising questions about the reasonableness and representativeness of the model's explanations. To address this limitation, future work could investigate the impact of using different definitions, such as definitions used by the Moderation API (Moderation API, 2023), to interact with ChatGPT.

The third type of limitation in our work relates to outputs and their meanings. While probability can provide some indication of the likelihood that a comment is classified as HOT, not all probabilities are informative. For instance, probabilities of 0.01, 0.05, or 0.95 only reflect the ChatGPT's assessment that a comment is highly likely or unlikely to be classified as HOT, but they do not provide much granularity about the model's degree of certainty. Furthermore, we have doubts about whether certain probabilities can accurately reflect the degree of HOT, particularly those in the "uncertain" range. For example, it remains unclear whether a comment with a probability of 0.6 is more likely to be classified as HOT than one with a probability of 0.4, as ChatGPT could interpret both probabilities as indicating that the comment may be perceived as HOT by certain individuals. Therefore, it may be necessary to develop additional methods to complement the use of probability in order to provide more nuanced insights into the degree of HOT. Possible avenues for future research could include exploring more inclusion of contextual information to improve the accuracy and interpretability of the model's outputs.

Two avenues of future work deserve further investigation. First, we plan to investigate the impact of MTurker annotators' demographics on the performance of ChatGPT. Specifically, we plan to test whether ChatGPT shows better agreement with annotations from certain demographics than others. It is also worth investigating whether ChatGPT can provide demographic-dependent answers as prior work has shown that certain demographic users may systematically differ in their annotations of HOT concepts. Overall, this line of research can help us to identify any biases or limitations in the model's ability to classify HOT comments across different demographic groups.

Second, given the rapid development of generative AI models, we plan to test the performance of other models in comparison to ChatGPT, such as the LLaMa model (Meta AI, 2023), using broader datasets related to harmful content. This will allow us to compare the strengths and limitations of different generative AI models and potentially identify promising approaches to improving the accuracy and interpretability of HOT classification.

# 6. Conclusions

This study investigates the potential of using generative AI models for annotating HOT comments and compares its results with those from MTurkers. Our findings show that ChatGPT exhibits approximately 80% accuracy and provides the same response more than 90% of the time in terms of reliability and consistency. It displays more consistent classification with MTurkers for non-HOT comments but less agreement for HOT comments. In particular, it shows significant disagreement with MTurkers when classifying "hateful" comments. When requesting probability outputs, we find that ChatGPT provides a greater number of results with extreme probabilities for HOT comments, and probabilities between 0.3 and 0.7 are less common. Next, our n-gram analysis shows that ChatGPT shows some conformity to the provided definitions, but its generalization ability for reasoning needs further study. Last, we find that different prompts can affect ChatGPT's performance, in particular requesting an explanation could probably receive more conservative outputs. In summary, we suggest that ChatGPT can be a useful tool for annotating large samples of content quickly and cheaply. However, to receive high-quality and useful annotations, researchers need to take care in defining the classes of HOT content and in designing the prompt they provide ChatGPT.

## Author contributions

- **Lingyao Li**: Conceptualization, Methodology, Data Curation, Formal Analysis, Writing - Original Draft, Writing - Review & Editing.
- **Lizhou Fan**: Conceptualization, Methodology, Formal Analysis, Writing - Original Draft, Writing - Review & Editing.
- **Shubham Atreja**: Conceptualization, Methodology, Writing - Original Draft, Writing - Review & Editing.
- **Libby Hemphill**: Conceptualization, Writing - Review & Editing, Project Administration, Resources, Supervision.

## Declaration of competing interest

The authors declare that they have no known competing interests or personal relationships that could have appeared to influence the work reported in this paper.

## Acknowledgement

We would like to express our thanks to Angela Gonzalez and Siqi Wu for their valuable viewpoints to this project. Their expertise and knowledge were instrumental in our analysis. Additionally, we appreciate their provision of the HOT dataset which enabled us to conduct our experiments. This material is based upon work supported by the National Science Foundation under grant no. 1928434.

# Appendices

## Appendix A.

Below are the details of prompts we provided to ChatGPT to implement our experiments.

**Prompt 1**

Provide the HOT concept and ask: "Do you think this comment is hateful? (1) Yes, (2) No." This is the same as how we asked MTurkers.

1. Do you think this comment is hateful? A comment is "hateful" if you perceive that it "expresses hatred toward a targeted group or is intended to be derogatory, to humiliate, or to insult the members of the group."
   a. Yes
   b. No

2. Do you think this comment is offensive? A comment is "offensive" if you perceive that it is "a hurtful, derogatory or obscene comment."
   a. Yes
   b. No

3. Do you think this comment is toxic? A comment is "toxic" if you perceive that it is "a rude, disrespectful, or unreasonable comment that is likely to make readers want to leave a discussion."
   a. Yes
   b. No

**Prompt 2**

Provide the HOT concept to the GPT model and only ask for a binary classification without further explanations.

1. I want you to perform a data annotation task. In your output, I want you to return either yes or no, depending on whether you think the following comment is hateful. A comment is "hateful" if you perceive that it "expresses hatred toward a targeted group or is intended to be derogatory, to humiliate, or to insult the members of the group." I want you to only respond with yes or no. Do not provide any other outputs or any explanation for your output.

2. I want you to perform a data annotation task. In your output, I want you to return either yes or no, depending on whether you think the following comment is offensive. A comment is "offensive" if you perceive that it is "a hurtful, derogatory or obscene comment." I want you to only respond with yes or no. Do not provide any other outputs or any explanation for your output.

3. I want you to perform a data annotation task. In your output, I want you to return either yes or no, depending on whether you think the comment is toxic. A comment is "toxic" if you perceive that it is "a rude, disrespectful, or unreasonable comment that is likely to make readers want to leave a discussion." I want you to only respond with yes or no. Do not provide any other outputs or any explanation for your output.

## Prompt 3

Provide the HOT concept to the GPT model and ask for a probability representing HOT without further explanations.

1. I want you to provide a probability score between 0 to 1 where the score represents the probability of the following comment being hateful. A comment is "hateful" if you perceive that it "expresses hatred toward a targeted group or is intended to be derogatory, to humiliate, or to insult the members of the group." A probability of 1 means that the comment is highly likely to be hateful. A probability of 0 means that the comment is highly unlikely to be hateful. Do not provide any other outputs or any explanation for your output.

2. I want you to provide a probability score between 0 to 1 where the score represents the probability of the following comment being offensive. A comment is "offensive" if you perceive that it is "a hurtful, derogatory or obscene comment." A probability of 1 means that the comment is highly likely to be offensive. A probability of 0 means that the comment is highly unlikely to be offensive. Do not provide any other outputs or any explanation for your output.

3. I want you to provide a probability score between 0 to 1 where the score represents the probability of the following comment being toxic. A comment is "toxic" if you perceive that it is "a rude, disrespectful, or unreasonable comment that is likely to make readers want to leave a discussion." A probability of 1 means that the comment is highly likely to be toxic. A probability of 0 means that the comment is highly unlikely to be toxic. Do not provide any other outputs or any explanation for your output.

## Prompt 4

Provide the HOT concept to the GPT model and only ask for a binary classification with further explanations.

1. I want you to perform a data annotation task. In your output, I want you to return either yes or no, depending on whether you think the following comment is hateful. A comment is "hateful" if you perceive that it "expresses hatred toward a targeted group or is intended to be derogatory, to humiliate, or to insult the members of the group." I want you to respond with yes or no in the first line and provide an explanation for your output in the second line.

2. I want you to perform a data annotation task. In your output, I want you to return either yes or no, depending on whether you think the following comment is offensive. A comment is "offensive" if you perceive that it is "a hurtful, derogatory or obscene comment." I want you to respond with yes or no in the first line and provide an explanation for your output in the second line.

3. I want you to perform a data annotation task. In your output, I want you to return either yes or no, depending on whether you think the comment is toxic. A comment is "toxic" if you perceive that it is "a rude, disrespectful, or unreasonable comment that is likely to make readers want to leave a discussion." I want you to respond with yes or no in the first line and provide an explanation for your output in the second line.

**Prompt 5**

Provide the HOT concept to the GPT model and ask for a probability representing HOT with further explanations.

1. I want you to provide a probability score between 0 to 1 where the score represents the probability of the following comment being hateful. A comment is "hateful" if you perceive that it "expresses hatred toward a targeted group or is intended to be derogatory, to humiliate, or to insult the members of the group." A probability of 1 means that the comment is highly likely to be hateful. A probability of 0 means that the comment is highly unlikely to be hateful. I want you to respond with a probabilistic score in the first line and provide an explanation for your score in the second line.

2. I want you to provide a probability score between 0 to 1 where the score represents the probability of the following comment being offensive. A comment is "offensive" if you perceive that it is "a hurtful, derogatory or obscene comment." A probability of 1 means that the comment is highly likely to be offensive. A probability of 0 means that the comment is highly unlikely to be offensive. I want you to respond with a probabilistic score in the first line and provide an explanation for your score in the second line.

3. I want you to provide a probability score between 0 to 1 where the score represents the probability of the following comment being toxic. A comment is "toxic" if you perceive that it is "a rude, disrespectful, or unreasonable comment that is likely to make readers want to leave a discussion." A probability of 1 means that the comment is highly likely to be toxic. A probability of 0 means that the comment is highly unlikely to be toxic. I want you to respond with a probabilistic score in the first line and provide an explanation for your score in the second line.

## Appendix B.

**Table B** displays the comparison results where the annotations from MTurkers were treated as a score. To ensure a fair assessment between ChatGPT's probability output and the Mturkers' score, we initially applied a specific rule to transform ChatGPT's probability output. This approach aimed to establish a standardized comparison between the two scoring methods, ensuring an "apple-to-apple" comparison. The probability output is based on Prompt 2.

- $0.0 \leq prob < 0.1 \Leftrightarrow 5$ "False" and 0 "True" from MTurkers (score = 0.0)
- $0.1 \leq prob < 0.3 \Leftrightarrow 4$ "False" and 1 "True" from MTurkers (score = 0.2)
- $0.3 \leq prob < 0.5 \Leftrightarrow 3$ "False" and 2 "True" from MTurkers (score = 0.4)
- $0.5 \leq prob < 0.7 \Leftrightarrow 2$ "False" and 3 "True" from MTurkers (score = 0.6)
- $0.7 \leq prob < 0.9 \Leftrightarrow 1$ "False" and 4 "True" from MTurkers (score = 0.8)
- $0.9 \leq prob \leq 1.0 \Leftrightarrow 0$ "False" and 5 "True" from MTurkers (score = 1.0)

**Table B**. Comparison between MTurker annotation and ChatGPT output on Prompt 2.

| Category | Score | Support | Precision | Recall | F1-score | Accuracy |
|---|---|---|---|---|---|---|
| Hateful | 0.0 | 2219 | 0.94 | 0.36 | 0.52 | 0.36 |
|  | 0.2 | 567 | 0.19 | 0.53 | 0.28 |  |
|  | 0.4 | 276 | 0.17 | 0.15 | 0.16 |  |
|  | 0.6 | 182 | 0.11 | 0.07 | 0.09 |  |
|  | 0.8 | 142 | 0.12 | 0.42 | 0.18 |  |
|  | 1.0 | 77 | 0.14 | 0.35 | 0.20 |  |
| Offensive | 0.0 | 1599 | 0.89 | 0.35 | 0.50 | 0.35 |
|  | 0.2 | 607 | 0.23 | 0.57 | 0.33 |  |
|  | 0.4 | 399 | 0.13 | 0.06 | 0.08 |  |
|  | 0.6 | 319 | 0.16 | 0.07 | 0.10 |  |
|  | 0.8 | 325 | 0.20 | 0.38 | 0.26 |  |
|  | 1.0 | 214 | 0.31 | 0.56 | 0.39 |  |
| Toxic | 0.0 | 1542 | 0.92 | 0.13 | 0.22 | 0.19 |
|  | 0.2 | 695 | 0.18 | 0.29 | 0.23 |  |
|  | 0.4 | 426 | 0.12 | 0.06 | 0.08 |  |
|  | 0.6 | 359 | 0.07 | 0.03 | 0.05 |  |
|  | 0.8 | 267 | 0.06 | 0.19 | 0.09 |  |
|  | 1.0 | 171 | 0.16 | 0.87 | 0.28 |  |

## Appendix C.

**Table C** displays samples that provide insight into ChatGPT's classification of HOT content, to support the discussion in **Section 4.4**. In particular, **Table C** presents specific examples that were classified as "toxic" but not "hateful" or "offensive," "toxic" and "offensive" but not "hateful," and "offensive" but not "hateful" or "toxic."

**Table C**. Examples of ChatGPT's outputs based on Prompt 4.

| Comment | ChatGPT's Response by Concept | | |
|---|---|---|---|
|  | Hateful | Offensive | Toxic |
| So should you Mr. Graham!! You are a traitor!! | No.<br><br>The comment does not express hatred towards a targeted group. It is directed towards an individual and accuses them of being a traitor. While it may be insulting or derogatory towards the individual, it does not target a specific group of people based on their race, ethnicity, | Yes.<br><br>The comment contains derogatory language ("traitor") directed towards Mr. Graham, which can be perceived as hurtful and offensive. | Yes<br><br>The comment contains disrespectful language and personal attacks towards Mr. Graham, which can make readers feel uncomfortable and want to leave the discussion. Therefore, it can be considered toxic. |
| | religion, or other characteristic. | | |
| Dumb governor | No.<br><br>The comment is not directed towards a specific group and does not express hatred towards any particular group. It is a general insult towards a governor, but it does not target any specific group or community. | Yes<br><br>The comment is derogatory as it uses the word "dumb" to insult the governor. It is a negative and hurtful comment that is intended to belittle and demean the person being referred to. Therefore, it can be considered offensive. | Yes<br><br>The comment is disrespectful and uses a derogatory term to refer to the governor. It is likely to offend and provoke negative reactions from readers, making them want to leave the discussion. Therefore, it meets the criteria for a toxic comment. |
| Blatant lies show they have zero shame. | No.<br><br>The comment does not express hatred towards a targeted group or intend to be derogatory, humiliate, or insult any members of a group. It is a statement criticizing someone's behavior or actions. | No.<br><br>The comment is critical and accusatory, but it does not contain any explicit derogatory or obscene language. It is a statement of opinion rather than a personal attack. | Yes.<br><br>The comment contains disrespectful language and implies that the person being referred to has no shame. This type of language is likely to make readers feel uncomfortable and want to leave the discussion. |
| I am going to stop paying my vehicle and mortgage loan. I think the taxpayer should pay it. | No.<br><br>The comment does not express hatred towards a targeted group or intend to be derogatory, humiliate, or insult any members of a group. It is simply a statement about the author's personal financial decisions and opinion on who should pay for their loans. | No.<br><br>The comment is not offensive as it does not contain any hurtful, derogatory, or obscene language. It may be considered irresponsible or unethical, but it does not meet the criteria for being offensive. | Yes.<br><br>This comment is toxic because it is disrespectful and unreasonable. The commenter is suggesting that they will stop paying their own debts and instead expect taxpayers to foot the bill. This kind of attitude is likely to provoke negative reactions from other readers and could lead to a breakdown in civil discussion. |
| This is the biggest clown show I've ever seen. The world is laughing at us. | No.<br><br>The comment does not express hatred towards a targeted | No.<br><br>The comment is critical and negative, but it does not contain any | Yes.<br><br>The comment contains disrespectful language and is likely to make |
| | group. It is a criticism of a situation or event, and does not target any specific group of people. While it may be insulting or derogatory towards those responsible for the situation, it does not aim to humiliate or insult any particular group. | hurtful, derogatory, or obscene language. It expresses the author's opinion about a situation, but it does not attack or insult any individual or group. | readers want to leave a discussion. The use of the word "clown" is derogatory and the statement "the world is laughing at us" is inflammatory. |
| Look at his face though, he's so big and strong that he'll DEFINITELY beat it. /s | No.<br><br>The comment does not express hatred towards a targeted group or intend to be derogatory, humiliate, or insult any members of a group. The comment is a sarcastic remark about someone's physical appearance and their ability to beat something. The "/s" at the end of the comment indicates that it is meant to be taken as sarcasm or a joke. | Yes<br><br>The comment is offensive because it is making a derogatory remark about someone's physical appearance and implies that their size and strength are the only factors that matter in a situation. Additionally, the use of the word "beat" suggests violence, which is also inappropriate. | No<br><br>The comment does not contain any rude, disrespectful, or unreasonable language that is likely to make readers want to leave a discussion. The use of sarcasm at the end of the comment indicates that it is not meant to be taken seriously. |
| Just ban her she's the dumbest person in America | Yes.<br><br>The comment is derogatory and insulting towards a specific person, which is a form of hate speech. The use of the word "dumbest" is intended to humiliate and insult the person, which is not acceptable behavior. | Yes.<br><br>This comment is offensive because it is derogatory and hurtful towards the person mentioned. The use of the word "dumbest" is insulting and implies that the person is unintelligent. Additionally, the suggestion to ban someone based on their perceived intelligence is discriminatory and exclusionary. Overall, this comment is not constructive and serves only to attack and belittle the individual. | Yes.<br><br>This comment is toxic because it is rude, disrespectful, and unreasonable. It attacks a specific person and uses derogatory language, which is likely to make readers feel uncomfortable and want to leave the discussion. |

| Can't fix stupid, well I guess covid can. | Yes | Yes | Yes |
|---|---|---|---|
| | The comment is derogatory and insulting towards individuals who may have contracted COVID-19. The phrase "Can't fix stupid" implies that those who have contracted the virus are somehow stupid or deserving of their illness. The second part of the comment, "well I guess covid can," suggests that the speaker believes that COVID-19 is a punishment for stupidity. This type of language is harmful and contributes to the stigmatization of individuals who have contracted the virus. | The comment is offensive because it uses derogatory language to insult someone's intelligence and makes light of a serious global pandemic that has caused immense suffering and loss of life. It is insensitive and lacks empathy towards those who have been affected by COVID-19. | The comment is disrespectful and uses a derogatory term to refer to people. It also makes light of a serious situation (COVID-19 pandemic) which can be seen as insensitive and inappropriate. Overall, the comment is likely to offend and discourage readers from engaging in a discussion. |